\documentclass[conference]{IEEEtran}
\usepackage[T1]{fontenc}
\usepackage{cite}
\usepackage{amsmath,amssymb,amsfonts}
\usepackage{graphicx}
\usepackage{xcolor}
\usepackage{booktabs}
\usepackage{multirow}
\usepackage{url}
\usepackage[hidelinks]{hyperref}
\usepackage{siunitx}
\usepackage{tabularx}
\usepackage{todonotes}
\usepackage{tikz}
\usepackage{threeparttable}
\usetikzlibrary{arrows.meta,positioning,shapes.geometric}
\usepackage{float}

\title{Characterizing LLM Inference Energy-Performance Tradeoffs across  Workloads and GPU Scaling}

\author{
    \IEEEauthorblockN{Paul Joe Maliakel}
    \IEEEauthorblockA{Computational Sustainability Group \\
    TU Wien, Austria\\
    paul.maliakel@tuwien.ac.at}
    \and
    \IEEEauthorblockN{Shashikant Ilager}
    \IEEEauthorblockA{Multiscale Networked Systems Group\\
    University of Amsterdam, Netherlands\\
    s.s.ilager@uva.nl}
    \and
    \IEEEauthorblockN{Ivona Brandic}
    \IEEEauthorblockA{Computational Sustainability Group\\
    TU Wien, Austria\\
    ivona.brandic@tuwien.ac.at}
}

\begin{document}
\maketitle

\begin{abstract}
LLM inference exhibits substantial variability across queries and execution
phases, yet inference configurations are often applied uniformly. We present a
measurement-driven characterization of workload heterogeneity and
energy--performance behavior of LLM inference under GPU dynamic voltage and
frequency scaling (DVFS). We evaluate five decoder-only LLMs (1B--32B
parameters) across four NLP benchmarks using a controlled offline setup. We show that lightweight semantic features predict inference difficulty better
than input length, with 44.5\% of queries achieving comparable quality across
model sizes. At the hardware level, the decode phase dominates inference time
(77--91\%) and is largely insensitive to GPU frequency. Consequently, reducing
GPU frequency from 2842\,MHz to 180\,MHz achieves an average of 42\% energy
savings with only a 1--6\% latency increase. We further provide a use case with an upper-bound
analysis of the potential benefits of combining workload-aware model selection
with phase-aware DVFS, motivating future energy-efficient LLM inference
systems.
\end{abstract}

\begin{IEEEkeywords}
Energy efficiency, LLM inference, benchmarking, DVFS
\end{IEEEkeywords}

\section{Introduction}
\label{sec:introduction}

Large Language Models (LLMs) have become a dominant interface for general-purpose
AI, powering applications such as interactive assistants, search, and enterprise
automation \cite{chowdhery2022palm, bommasani2021foundation, zhao2025surveylargelanguagemodels, brown2020language,openai2023gpt4}. Despite rapid algorithmic progress, the cost of LLM inference in
terms of energy, latency, and infrastructure provisioning has emerged as a
first-order concern for sustainable AI systems
\cite{strubell2019energy,schwartz2019greenai}.

In production, LLM serving systems operate on heterogeneous hardware and must
handle diverse query streams. These range from short factual questions to long-context
reasoning prompts, with outputs varying from a few tokens to long-form generation
\cite{mlperf_inference,revel2024optimml,kwon2023vllm}. However, existing inference
systems typically apply uniform configurations across all queries, implicitly
assuming homogeneous workloads and uniform hardware sensitivity. This motivates a
systematic analysis of how workload characteristics and hardware configurations
jointly affect inference performance and energy cost.

Prior work on efficient LLM inference has focused on system-level optimizations
such as batching, KV-cache management, and scheduling
\cite{kwon2023vllm, dao2022flashattention,zheng2022orca}. While effective for
throughput and latency, these approaches largely rely on coarse-grained workload
descriptors, such as input length or token count. Consequently, the interaction
between semantic query characteristics, execution phases, and low-level hardware
controls such as GPU frequency scaling remains insufficiently understood from an
energy-efficiency perspective.

At the workload level, inference cost is often assumed to scale with input length
\cite{mlperf_inference}. In practice, this assumption frequently breaks down. Short queries may be semantically difficult due to dense entity usage or
implicit reasoning requirements \cite{talmor2019commonsenseqa,geva2021strategyqa,
valmeekam2022planbench}, while long queries may involve comparatively simple
retrieval or summarization \cite{bai2023longbench,zhong2021qmsum,
kovcisky2018narrativeqa}. These observations motivate workload characterization
based on linguistic and semantic properties beyond surface-level token counts
\cite{ethayarajh2022understanding}.


At the hardware level, the energy consumption and latency of LLM inference are strongly influenced by accelerator configuration choices, particularly GPU Dynamic Voltage and Frequency Scaling (DVFS). Existing systems typically rely on static or coarse-grained frequency settings during inference, without adapting to phase-level variations in computational demand \cite{10540202}. However, decoder-only inference exhibits a pronounced structural asymmetry: a compute-bound \emph{prefill} phase that processes input tokens in parallel, followed by a memory-bound autoregressive \emph{decode} phase characterized by repeated accesses to model weights and the growing KV cache \cite{10609649, dao2022flashattention, shoeybi2019megatron}. This heterogeneity suggests that applying a uniform GPU frequency across both phases may be inherently suboptimal for energy efficiency and performance.


Accordingly, this paper presents a measurement-driven benchmarking study that
jointly characterizes workload-level query heterogeneity and hardware-level
energy--performance behavior under GPU DVFS. Using a controlled offline
replay-based setup, we evaluate five decoder-only LLMs (1B--32B parameters) across
four representative NLP benchmarks spanning classification and generation tasks: BoolQ, HellaSwag, TruthfulQA, and NarrativeQA
\cite{clark2019boolq,zellers2019hellaswag,lin2022truthfulqa,kovcisky2018narrativeqa}.
We perform systematic GPU frequency sweeps and measure latency and energy at both
end-to-end and phase-level granularity.
 
Our goal is not to propose a new serving system or scheduling algorithm, but to
provide empirical insights into where and why energy is consumed during LLM
inference, and which workload and hardware characteristics create opportunities
for energy savings. These characterizations form a foundation for workload-aware
and phase-aware optimization strategies in future LLM serving systems.


Our measurements show that semantic query characteristics provide stronger
signals of workload heterogeneity than input length alone, and that the decode
phase dominates inference time and energy consumption while remaining largely
insensitive to GPU frequency scaling. A case study further demonstrates how
combining workload-aware model selection with phase-aware frequency configuration
can substantially improve inference energy efficiency without sacrificing
latency.

In summary, this paper makes the following contributions:
\begin{itemize}
  \item \textbf{Workload characterization:} We analyze lexical, syntactic, and
  semantic query features and show that semantic properties (e.g., entity
  density) explain inference difficulty better than input length
  (Section~\ref{sec:workload}).
  \item \textbf{DVFS characterization:} We benchmark GPU DVFS across five models
  and four workloads, revealing a consistent phase-level asymmetry in which the
  decode phase is comparatively frequency-insensitive
  (Section~\ref{sec:dvfs_characterization}).
  \item \textbf{Implications via case study:} We demonstrate how workload-aware
  model selection and phase-aware DVFS can be combined to improve energy
  efficiency while maintaining quality
  (Section~\ref{sec:casestudy}).
\end{itemize}

Our results provide a benchmarking foundation for energy-efficient LLM
inference policies grounded in workload heterogeneity and phase-aware hardware
behavior rather than fixed conservative defaults.

\section{Background}
\label{sec:background}

\subsection{Transformer-based LLM Inference}
\label{sec:background:transformers}

Transformer decoders generate tokens autoregressively using stacked attention
and feed-forward layers \cite{vaswani2017attention}. Modern open LLMs span a wide
range of sizes and architectures, including GPT-style decoders
\cite{radford2019gpt2}, OPT-style baselines \cite{zhang2022opt}, LLaMA-family
models \cite{touvron2023llama}, and recent instruction-tuned families such as
Qwen and Mistral \cite{ai2024qwen2_5,jiang2023mistral}. Inference efficiency is
strongly influenced by memory traffic from model weights and the KV cache, as
well as attention implementations that reduce HBM (High Bandwidth Memory) accesses
\cite{dao2022flashattention}.

\subsection{Prefill and Decode Phases}
\label{sec:background:phases}

Decoder-only inference consists of two phases. During \textbf{prefill}, the model
processes the entire input sequence to compute hidden states and populate the KV
cache, typically exhibiting high arithmetic intensity. During \textbf{decode},
output tokens are generated sequentially, requiring repeated access to model
weights and the growing KV cache, which often makes decode memory-dominated
\cite{kwon2023vllm,dao2022flashattention,shoeybi2019megatron}. Serving systems
optimize these phases through batching and KV-cache management, as implemented in
high-throughput engines such as vLLM \cite{kwon2023vllm}.

\subsection{GPU DVFS and Energy--Performance Tradeoffs}
\label{sec:background:dvfs}

Dynamic Voltage and Frequency Scaling (DVFS) adjusts GPU operating frequency to
trade energy for performance. Compute-bound kernels typically scale with
frequency, while memory-bound kernels are limited by bandwidth and show weaker
scaling \cite{liu2025greenllmsloawaredynamicfrequency}. This distinction motivates phase-aware analysis for LLM inference.
Prior work has explored energy--performance tradeoffs using DVFS, power limits,
and configuration search \cite{you2023zeus,chung2024perseus,
nabavinejad2021batchsizer,revel2024optimml}. In this study, we focus on DVFS
characterization for LLM inference and measure GPU power using NVML telemetry.

\subsection{Workload Heterogeneity in LLM Inference}
\label{sec:background:workload}

Benchmarking LLM inference requires representative workloads and standardized
metrics. MLPerf Inference highlights the importance of end-to-end comparability
across systems \cite{mlperf_inference}. LLM workloads further exhibit substantial
heterogeneity: query streams vary widely in semantic difficulty, and generation
tasks increasingly shift execution toward decode-dominated regimes. We use
established NLP benchmarks that capture complementary difficulty profiles,
including BoolQ and HellaSwag (classification), TruthfulQA (misconception-prone
factuality), and NarrativeQA (long-context comprehension)
\cite{liu2025greenllmsloawaredynamicfrequency,zellers2019hellaswag,lin2022truthfulqa, kovcisky2018narrativeqa}.

\section{Related Work}
\label{sec:related_work}

Prior work on efficient LLM inference spans benchmarking methodologies,
system-level optimizations, adaptive inference, and energy-aware execution.
The growing energy and environmental footprint of deep learning has motivated
measurement-driven approaches and ``Green AI'' principles
\cite{strubell2019energy,schwartz2019greenai}, while benchmark suites such as
MLPerf Inference emphasize standardized, end-to-end evaluation of inference
performance across platforms and configurations \cite{mlperf_inference}.
Efficient LLM serving systems improve throughput and latency through batching
and KV-cache management, as shown by vLLM and its PagedAttention mechanism
\cite{kwon2023vllm}, and through optimized attention kernels such as
FlashAttention that reduce HBM traffic \cite{dao2022flashattention}. Other
systems and toolchains explore optimized and distributed inference runtimes,
including TensorRT-based stacks and large-scale distributed frameworks
\cite{shoeybi2019megatron}. A complementary line of work reduces
decode cost through speculative execution or multi-head decoding
\cite{levine2023stagedspec,cai2024medusa}, post-training quantization techniques
such as SmoothQuant \cite{xiao2023smoothquant}, or adaptive inference strategies
including model cascades and cost-aware routing
\cite{chen2023frugalgpt,teerapittayanon2016branchynet}. Separately, DVFS and
power-aware execution have been explored as knobs for improving energy efficiency
in GPU-based ML workloads, using configuration search or power--performance
tradeoff analysis \cite{10540202, you2023zeus,chung2024perseus,
nabavinejad2021batchsizer,revel2024optimml}. In contrast to these efforts, our
work provides a measurement-driven, phase-aware DVFS characterization for LLM
inference across model sizes and workloads, and explicitly connects hardware
sensitivity to workload and query heterogeneity.

\section{Design and Methodology}
\label{sec:experimental_setup}

This section describes the design and methodology of our study, and provides details about the hardware platform, models, datasets, and evaluation
metrics used in our benchmarking study. All experiments use an offline,
replay-based inference setup to ensure reproducibility.

\subsection{Design}

\begin{figure}[t]
  \centering
  \includegraphics[width=\columnwidth]{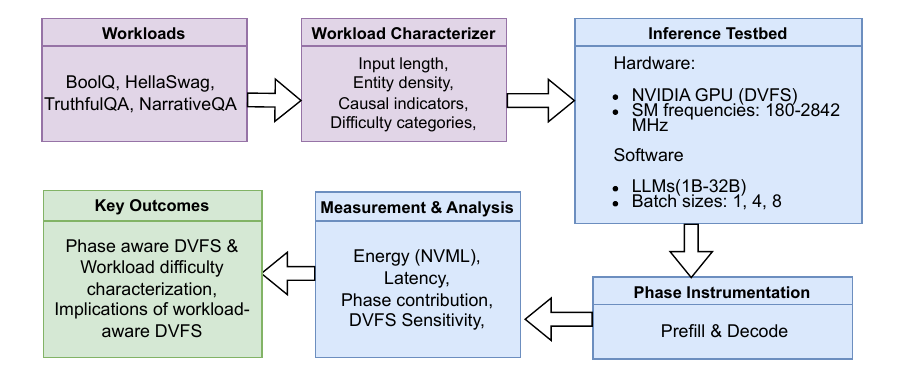}
  \caption{Study workflow.}
  \label{fig:workflow}
\end{figure}

Figure~\ref{fig:workflow} shows the benchmarking workflow. We conduct an offline,
measurement-driven study that evaluates representative NLP workloads using a
DVFS-controlled GPU testbed. Queries are first characterized using lightweight
semantic features, after which inference is executed while varying model size,
batch size, and SM frequency under a fixed decoding configuration. Inference is
instrumented to separate prefill and decode phases, and GPU power and latency are
recorded via NVML. These measurements are aggregated to derive phase-aware DVFS
behavior, workload difficulty characterization, and implications of workload aware DVFS.







\subsection{Testbed and Measurement Infrastructure}
All measurements are performed on a single NVIDIA RTX~PRO~6000 (Blackwell) GPU with 96\,GB memory, which supports explicit dynamic voltage and frequency scaling (DVFS). The streaming multiprocessor (SM) clock is fixed via NVIDIA’s management interface for the duration of each experiment. We evaluate seven SM frequency levels: 180, 487, 960, 1500, 2000, 2505, and 2842\,MHz, while
keeping GPU memory frequency at its default setting to isolate SM scaling
effects. Unless stated otherwise, we use the maximum supported SM frequency (2842 MHz) as the baseline configuration for all comparisons. Power consumption is measured using NVIDIA Management Library (NVML) telemetry via \texttt{nvidia-smi}, sampled at 10\,ms and integrated to compute per-request energy in joules. All timing measurements use \texttt{torch.cuda.synchronize()} to ensure GPU 
  operations complete before recording timestamps. Each configuration is repeated three times, and we report mean values. Experiments follow an offline replay-based setup on an otherwise idle system, using batch sizes of 1, 4, and 8 to ensure controlled and repeatable energy and latency measurements.

\subsection{Models}
\label{sec:models}

We evaluate five decoder-only large language models spanning 1B to 32B parameters,
representing widely used open-source architectures
(Table~\ref{tab:models_datasets}). All models use publicly released pre-trained
weights without fine-tuning. Inference is performed in FP16 precision with
identical decoding configurations across models to ensure comparability.

For generation tasks, all inference experiments use greedy decoding
(\texttt{do\_sample=False}, \texttt{temperature=0}), with a maximum of 100 generated
tokens and early stopping on the end-of-sequence token. For classification tasks
(BoolQ and HellaSwag), we use log-likelihood evaluation by comparing answer option
probabilities without token generation, ensuring deterministic and comparable
energy measurements across models.

 \begin{table}[t]
\centering
\caption{Models and Datasets Used in Evaluation}
\label{tab:models_datasets}
\footnotesize
\setlength{\tabcolsep}{2.5pt}
\begin{tabular}{lcc|lccc}
\toprule
\multicolumn{3}{c|}{\textbf{Models}} & \multicolumn{4}{c}{\textbf{Datasets (Avg. Tokens)}} \\
\cmidrule(lr){1-3} \cmidrule(lr){4-7}
\textbf{Model} & \textbf{Params} & \textbf{Arch.} &
\textbf{Dataset} & \textbf{Task} & \textbf{Input} & \textbf{Output} \\
\midrule
Llama-3.2-1B  & 1B  & Decoder-only & BoolQ        & Class. & 100 & LL   \\
Llama-3.2-3B  & 3B  & Decoder-only & HellaSwag    & Class. & 166 & LL   \\
Llama-3.1-8B  & 8B  & Decoder-only & NarrativeQA  & Gen.   & 336 & 100 \\
Qwen2.5-14B   & 14B & Decoder-only & TruthfulQA   & Gen.   & 13  & 100 \\
Qwen2.5-32B   & 32B & Decoder-only &              &        &     &     \\
\bottomrule
\end{tabular}
\par\vspace{2pt}
{\footnotesize \textit{LL = log-likelihood evaluation (no token generation).}}    
\end{table}

\subsection{Datasets and Metrics}
\label{sec:datasets_metrics}

We evaluate inference workloads using four widely used NLP benchmarks: BoolQ,
HellaSwag, NarrativeQA, and the generative variant of TruthfulQA
(TruthfulQA\_GEN, referred to as TruthfulQA in the remainder of the paper). These datasets span both classification
and generative tasks and exhibit substantial variation in input and output
lengths. BoolQ and HellaSwag represent classification-style workloads with short
outputs, while NarrativeQA and TruthfulQA require free-form generation with
longer decode phases. Each dataset is evaluated across all five models using identical decoding configurations, unless explicitly stated otherwise. Table~\ref{tab:models_datasets} summarizes key dataset
characteristics.

We evaluate each inference configuration using metrics that capture both
efficiency and output quality. \textbf{Energy} is measured as total GPU energy
consumption in joules by integrating power over inference duration.
\textbf{Latency} is measured as end-to-end inference time from input submission
to completion of output generation. \textbf{Quality} is measured using accuracy
for classification tasks and ROUGE-L for generative tasks. We additionally report
the \textbf{Energy--Delay Product (EDP= Energy × Latency)} to capture the tradeoff between energy
consumption and latency.

\section{Workload Characterization}
\label{sec:workload}

\subsection{Motivation and Scope}
\label{sec:workload:motivation}

Understanding input query characteristics is essential for designing
energy-efficient LLM inference systems, as queries differ widely in their
computational demands and response to model scaling. A key motivation of this
work is that not all queries require the same model capacity: simple factual
queries can often be handled effectively by smaller models, while complex
reasoning queries benefit from larger models.

Building on the benchmark datasets described in Section~\ref{sec:datasets_metrics}, we
focus on observable properties of input queries and generated outputs that
influence inference behavior and energy consumption. We analyze lightweight
structural, linguistic, and semantic features to distinguish easy and hard
queries, validate that these features capture semantic difficulty beyond
surface-level properties such as input length, and examine how query difficulty
interacts with model scale. All features are extracted prior to inference and
incur negligible runtime overhead. We do not claim to propose an optimal or deployable difficulty predictor; rather, we demonstrate that lightweight linguistic and semantic features provide stronger explanatory and predictive signals of inference difficulty than length-based heuristics.

\subsection{Input Length and Structural Properties}
\label{sec:workload:length}

\begin{table}[t]
\centering
\caption{Input Length Statistics (Tokens)}
\label{tab:input_length}
\footnotesize
\setlength{\tabcolsep}{4pt}
\begin{tabular}{lccccc}
\toprule
\textbf{Dataset} & \textbf{Mean} & \textbf{Std} & \textbf{Min} & \textbf{Max} & \textbf{Range} \\
\midrule
TruthfulQA   & 12.6  & 5.7  & 5   & 52  & 10.4$\times$ \\
BoolQ        & 102.9 & 46.0 & 24  & 294 & 12.2$\times$ \\
HellaSwag    & 163.8 & 56.0 & 49  & 265 & 5.4$\times$  \\
NarrativeQA  & 339.1 & 34.3 & 208 & 396 & 1.9$\times$  \\
\bottomrule
\end{tabular}
\end{table}
Input query length varies substantially across datasets, reflecting different
workload profiles. Table~\ref{tab:input_length} summarizes token length
statistics across 3,817 queries (1,000 per dataset, 817 for TruthfulQA).

Mean input length spans a 26.8$\times$ range across datasets, from short factual
queries in TruthfulQA to long-context inputs in NarrativeQA. Short-context
queries incur minimal prefill cost but rely on parametric knowledge, while
long-context queries shift computation toward prefill and enable in-context
retrieval. However, despite this large variation, input length alone does not
capture differences in linguistic structure or reasoning complexity. We
therefore analyze higher-level linguistic and semantic features that reflect
query difficulty beyond token counts.

\subsection{Linguistic and Semantic Complexity}
\label{sec:workload:complexity}

We characterize each query using lexical, syntactic, and semantic features. We intentionally focus on lightweight, interpretable features that can be extracted online with negligible overhead, rather than learned difficulty predictors that require additional inference cost. To identify the most informative ones, we compute correlations between each feature and output quality across all models, retaining features with $|r| > 0.10$ and low redundancy (pairwise correlation $< 0.7$). This yields five key features capturing complementary aspects of query complexity(Table~\ref{tab:complexity_features}):

\begin{itemize}
    \item \textbf{Complexity Score} (0--1): Weighted combination of normalized token entropy, unique token ratio, entity density, and average sentence length.
    \item \textbf{Reasoning Complexity} (0--1): Density of causal and comparison markers (e.g., ``because'', ``therefore'', ``however''), normalized by word count.
      \item \textbf{Entity Density} (0--1): Ratio of named entities to total tokens,                                                                        
  computed using spaCy's \texttt{en\_core\_web\_sm} NER model (PERSON, ORG, GPE, LOC types).    
    \item \textbf{Token Entropy} (bits): Shannon entropy of the token distribution,  $H = -\sum_i p_i \log_2 p_i$, where $p_i$ is the relative frequency 
  of token $i$
  \item \textbf{Causal Question Score} (0--100\%): Percentage of causal question words 
  (``why'', ``how'', ``explain'', ``justify'', ``prove'') relative to total question count. 
\end{itemize}


\begin{table}[t]
  \centering
  \caption{Input Complexity Features by Dataset (Mean Values)}
  \label{tab:complexity_features}
  \footnotesize
  \setlength{\tabcolsep}{4pt}
  \begin{tabular}{lcccc}
    \toprule
    \textbf{Feature} & \textbf{BoolQ} & \textbf{HellaSwag} & \textbf{TruthfulQA} & \textbf{NarrativeQA} \\
    \midrule
    Complexity Score     & 0.54 & 0.47 & 0.53 & 0.56 \\
    Reasoning Complexity & 0.06 & 0.11 & 0.07 & 0.12 \\
    Entity Density       & 0.20 & 0.12 & \textbf{0.34} & 0.18 \\
    Token Entropy        & 5.82 & 6.31 & 3.50 & \textbf{7.16} \\
    Causal Questions (\%) & 2.4  & 4.4  & 10.2 & \textbf{33.6} \\
    \bottomrule
  \end{tabular}
\end{table}

Across datasets, distinct semantic profiles emerge. TruthfulQA exhibits the highest       
  entity density (0.34), reflecting reliance on parametric factual recall, while BoolQ shows substantially
lower entity density (below 20\%). NarrativeQA displays higher token entropy
(typically exceeding 4.5 bits/token) and a larger fraction of causal questions
(approximately 30--35\%), indicating multi-step reasoning over provided context.
In contrast, HellaSwag and NarrativeQA show higher reasoning complexity scores
than BoolQ, consistent with their focus on commonsense inference and narrative
continuation rather than binary factual verification.

\begin{table}[t]
\centering
\caption{Causal Question Distribution by Dataset}
\label{tab:causal_questions}
\footnotesize
\setlength{\tabcolsep}{4pt}
\begin{tabular}{lcc}
\toprule
\textbf{Dataset} & \textbf{Causal Questions (\%)} & \textbf{Dominant Query Type} \\
\midrule
BoolQ        & 2.4  & Factual verification \\
HellaSwag    & 4.4  & Sequence prediction \\
TruthfulQA   & 10.2 & Factual and causal \\
NarrativeQA  & 33.6 & Comprehension and causal \\
\bottomrule
\end{tabular}
\end{table}

NarrativeQA contains substantially more causal questions (33.6\%), requiring multi-step reasoning across narrative context, while BoolQ primarily contains factual verification questions (2.4\% causal).
To assess whether these linguistic and semantic features provide information
beyond input length, we next analyze their independence from basic structural
properties and validate their explanatory power.


\subsection{Feature Independence and Validation}
\label{sec:workload:validation}

A key question is whether the proposed workload features capture semantic
difficulty beyond input length alone. We evaluate whether these features provide
information independent of length, rather than acting as proxies for length.

\subsubsection{Length Independence Analysis}

Table~\ref{tab:feature_length} reports correlations between each feature and input
length. While token entropy is strongly correlated with length ($r = +0.88$),
most semantic features, including entity density, causal question score, and
reasoning complexity, exhibit weak correlations with length. This indicates that
these features capture properties beyond input size.

\begin{figure}[t]
    \centering
    \includegraphics[width=0.85\columnwidth]{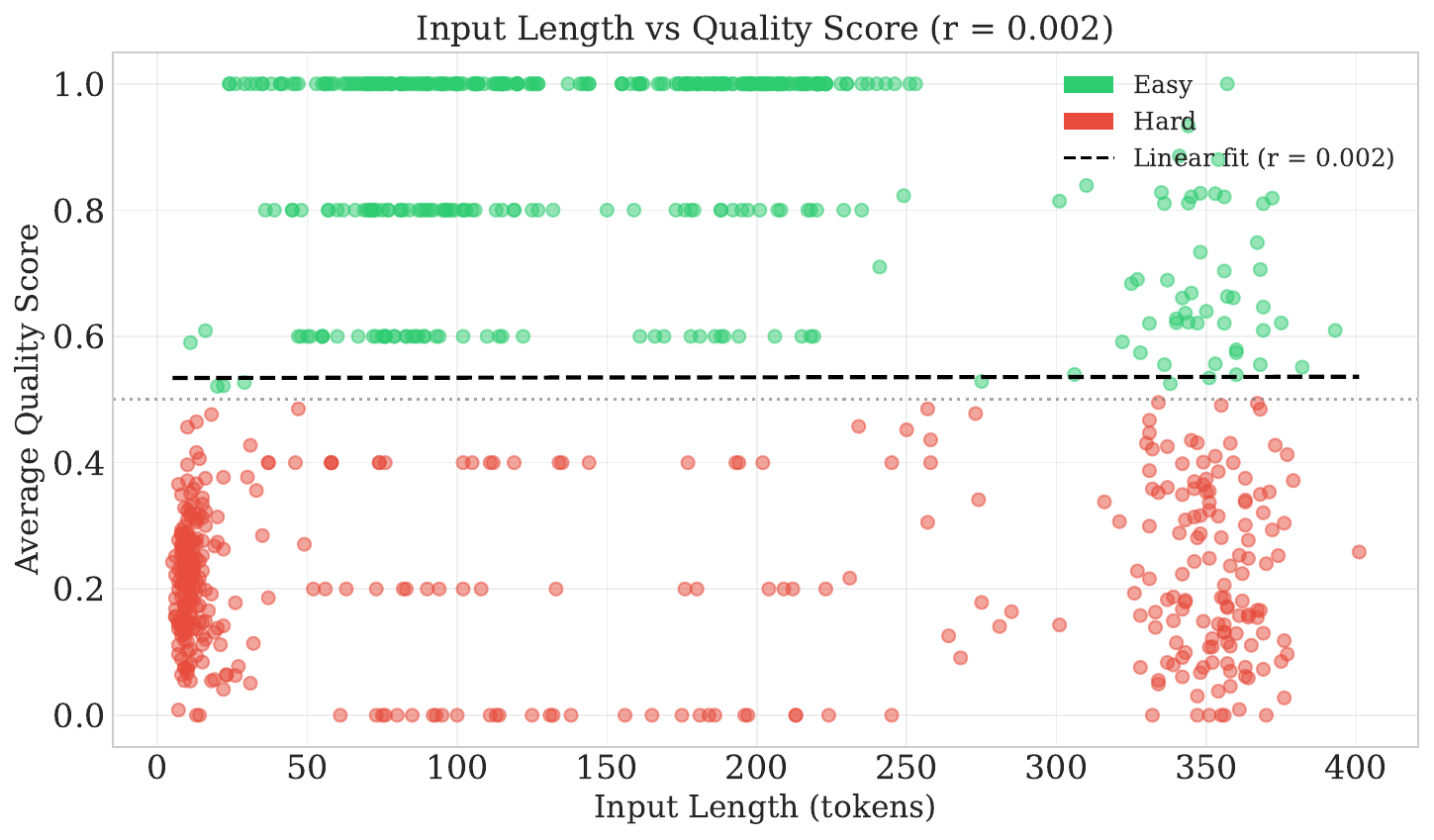}
    \caption{Input length vs quality score. The near-zero correlation ($r = 0.002$)                      
  demonstrates that length alone cannot predict query difficulty. Easy/hard labels  
   indicate whether normalized mean quality across models exceeds 0.5.}
    \label{fig:length_quality}
\end{figure}

Notably, input length itself shows near-zero correlation with output quality
($r = 0.002$), as shown in Figure~\ref{fig:length_quality}. This demonstrates
that length alone is not a reliable indicator of query difficulty. In contrast,
semantic features retain meaningful associations with quality even after
controlling for length through partial correlation analysis, confirming that
they capture semantic workload properties independently of input length.

\begin{table}[t]
\centering
\caption{Feature Independence from Input Length}
\label{tab:feature_length}
\begin{tabular}{lcc}
\toprule
\textbf{Feature} & \textbf{Corr. with Length} & \textbf{Independent?} \\
\midrule
Entity Density & $r = -0.44$ & \checkmark Yes \\
Causal Question Score & $r = +0.31$ & \checkmark Yes \\
Reasoning Complexity & $r = +0.19$ & \checkmark Yes \\
Token Entropy & $r = +0.88$ & $\times$ No \\
Complexity Score & $r = +0.16$ & \checkmark Yes \\
\midrule
\textit{Length $\rightarrow$ Quality} & \multicolumn{2}{c}{$r = 0.002$ (near zero)} \\
\bottomrule
\end{tabular}
\end{table}


\subsubsection{Ablation Study}
We define query difficulty using a binary label. Queries are
labeled as \textit{easy} if the normalized average quality score across all
five models exceeds the dataset median, and \textit{hard} otherwise. Quality
scores are min-max normalized within each dataset prior to averaging, ensuring
comparable scales between accuracy (classification tasks) and ROUGE-L
(generation tasks). This yields a balanced split of 49\% easy and 51\% hard
queries. 

We train a logistic regression classifier with L2 regularization
(C=1.0) using 5-fold stratified cross-validation; all features are standardized
to zero mean and unit variance before training. The baseline uses a simple
length threshold of 150 tokens, which achieves near-random performance (51.1\%). We include a simple length-only heuristic for comparison.

Table~\ref{tab:ablation} summarizes classification accuracy for different feature
sets. Adding semantic features improves accuracy by 17.5 percentage points over
the length-only baseline. Notably, semantic features alone achieve the same
accuracy as the combined feature set, confirming that input length provides no
additional predictive value beyond semantic workload characteristics.

\begin{table}[t]
\centering
\caption{Feature Ablation: Difficulty Classification Accuracy}
\label{tab:ablation}
\begin{tabular}{lc}
\toprule
\textbf{Feature Set} & \textbf{Accuracy (5-fold CV)} \\
\midrule
Length only ($>$150 tokens) & 51.1\% \\
+ Entity density & 66.6\% \\
+ Causal question score & 68.4\% \\
Features only (no length) & \textbf{68.6\%} \\
\bottomrule
\end{tabular}
\end{table}


\subsection{Query Difficulty Categories}
\label{sec:workload:difficulty}

Based on the validated features, we develop a model-size-aware difficulty classification by analyzing per-query performance across five models (1B to 32B parameters).

\subsubsection{Baseline Quality by Dataset}
Table~\ref{tab:quality_by_dataset} shows that model scaling consistently improves
quality, with average scores increasing from 0.423 (1B) to 0.596 (32B). Dataset
difficulty varies substantially, with BoolQ and HellaSwag achieving higher
accuracy than TruthfulQA and NarrativeQA, indicating different underlying
difficulty characteristics.

\begin{table}[t]
\centering
\caption{Quality Scores by Model and Dataset}
\label{tab:quality_by_dataset}
\begin{tabular}{lccccc|c}
\toprule
\textbf{Dataset} & \textbf{1B} & \textbf{3B} & \textbf{8B} & \textbf{14B} & \textbf{32B} & \textbf{Avg} \\
\midrule
BoolQ & 0.685 & 0.785 & 0.855 & 0.785 & 0.815 & 0.785 \\
HellaSwag & 0.640 & 0.755 & 0.805 & 0.830 & 0.860 & 0.778 \\
TruthfulQA & 0.208 & 0.211 & 0.207 & 0.243 & 0.252 & 0.224 \\
NarrativeQA & 0.161 & 0.306 & 0.368 & 0.474 & 0.455 & 0.353 \\
\midrule
\textbf{Model Avg} & \textbf{0.423} & \textbf{0.514} & \textbf{0.559} & \textbf{0.583} & \textbf{0.596} & \textbf{0.535} \\
\bottomrule
\end{tabular}
\vspace{1mm}
{\footnotesize
\textit{Note: Classification reports accuracy; generation reports ROUGE-L.}
}

\end{table}

\subsubsection{Feature-Quality Correlations by Model Size}

Table~\ref{tab:feature_model_corr} presents correlation coefficients between input features and quality scores for each model size tier.

\begin{table}[t]
\centering
\caption{Feature-Quality Correlations by Model Size}
\label{tab:feature_model_corr}
\begin{tabular}{lccccc}
\toprule
\textbf{Feature} & \textbf{1B} & \textbf{3B} & \textbf{8B} & \textbf{14B} & \textbf{32B} \\
\midrule
Entity Density & $-$0.20 & $-$0.29 & $-$0.32 & $-$0.31 & $-$0.32 \\
Causal Question & $-$0.16 & $-$0.17 & $-$0.18 & $-$0.13 & $-$0.18 \\
Token Entropy & +0.11 & +0.22 & +0.29 & +0.33 & +0.31 \\
\bottomrule
\end{tabular}
\end{table}

The analysis reveals two primary difficulty predictors. Entity density and causal question score emerge as the most consistent predictors
of difficulty across all model sizes, exhibiting stable negative correlations with
quality (typically $r \in [-0.18, -0.30]$ for entity density and
$r \in [-0.15, -0.25]$ for causal questions). In contrast, other features show
weaker or less consistent correlations (generally $|r| < 0.10$) and are excluded
to maintain a lightweight difficulty model.

\textbf{Feature Selection Rationale.} Of the five candidate features analyzed, we retain only \textbf{entity density} and \textbf{causal question score} as primary difficulty predictors for classification. The other features are excluded for the following reasons:
\begin{itemize}
    \item \textbf{Token Entropy}: Highly correlated with input length ($r = +0.88$, Table~\ref{tab:feature_length}), making it redundant with a surface-level property we have shown to be non-predictive of difficulty.
    \item \textbf{Reasoning Complexity}: Near-zero correlation with quality ($r = -0.03$), providing negligible predictive power despite theoretical relevance.
    \item \textbf{Complexity Score}: Weak correlation with quality ($r = -0.05$), insufficient to justify inclusion in a classifier.
\end{itemize}
Entity density and causal question score satisfy both criteria for effective difficulty prediction: they are independent from input length ($|r| < 0.5$) and show consistent negative correlations with quality across all model sizes.

\subsubsection{Scaling Pattern Analysis}

By analyzing performance trajectories across model sizes, we categorize queries into four scaling patterns. Quality scores are min-max normalized within each dataset before classification, ensuring that accuracy (0--1) and ROUGE-L scores are on comparable scales. Table~\ref{tab:scaling_patterns} summarizes the resulting distribution.

\begin{table}[t]
\centering
\caption{Query Scaling Patterns Across Model Sizes}
\label{tab:scaling_patterns}
\footnotesize
\setlength{\tabcolsep}{4pt}
\begin{tabular}{lcc}
\toprule
\textbf{Pattern} & \textbf{\%} & \textbf{Mean Feature Profile} \\
\midrule
Always Easy    & 44.5 & Entity=0.17,\; Causal=0.05,\; Entropy=6.13 \\
Scaling Helps  & 15.5 & Entity=0.18,\; Causal=0.13,\; Entropy=6.32 \\
Always Hard    & 32.6 & Entity=0.27,\; Causal=0.22,\; Entropy=4.94 \\
Inconsistent   & 7.4  & Architecture-dependent \\
\bottomrule
\end{tabular}
\end{table}

\begin{itemize}
    \item \textbf{44.5\% of queries are easy for all models}: These can be safely routed to small models (1--3B) for maximum energy efficiency.
    \item \textbf{15.5\% benefit from scaling}: These fail on small models but succeed on 8B+, representing optimal routing opportunities.
    \item \textbf{32.6\% are hard for all models}: Even 32B models struggle; routing to larger models provides marginal benefit at significant energy cost.
\end{itemize}

\subsubsection{Classification Validation}

We validate the feature-based difficulty classification against empirical model performance. Queries are classified as \textit{easy} if entity density $<$ 0.20 and causal question score $<$ 0.05, and \textit{hard} otherwise. This yields 406 easy (50.8\%) and 394 hard (49.2\%) queries. Table~\ref{tab:validation_quality} shows that all five models consistently achieve higher quality scores on easy queries, with an average gap of +0.256 (63\% relative improvement), validating the feature-based classification.

\begin{table}[t]
\centering
\caption{Classification Validation: Quality by Difficulty Category}
\label{tab:validation_quality}
\begin{tabular}{lcccc}
\toprule
\textbf{Model} & \textbf{Easy} & \textbf{Hard} & \textbf{Gap} & \textbf{Valid?} \\
\midrule
Llama-3.2-1B & 0.516 & 0.328 & +0.188 & \checkmark \\
Llama-3.2-3B & 0.637 & 0.388 & +0.250 & \checkmark \\
Llama-3.1-8B & 0.695 & 0.419 & +0.276 & \checkmark \\
Qwen2.5-14B & 0.713 & 0.450 & +0.263 & \checkmark \\
Qwen2.5-32B & 0.745 & 0.441 & +0.304 & \checkmark \\
\midrule
\textbf{Average} & \textbf{0.661} & \textbf{0.405} & \textbf{+0.256} & \checkmark \\
\bottomrule
\end{tabular}
\end{table}


\subsection{Main Workload Characterization Insights}
\label{sec:workload:summary}

Our workload characterization yields the following insights for energy-efficient
LLM inference:

\begin{enumerate}
    \item \textbf{Input length is a weak difficulty predictor} ($r = 0.002$),
    while semantic features such as entity density and causal question presence
    provide stronger signals.

    \item \textbf{Entity density is the dominant predictor} ($r = -0.29$),
    disproportionately challenging smaller models.

    \item \textbf{44.5\% of queries are easy across all model sizes}, enabling
    substantial energy savings via routing to smaller models.

    \item \textbf{32.6\% of queries are hard across all models}, where scaling
    yields diminishing quality returns.
\end{enumerate}

These properties shape inference energy consumption by influencing model-scale
sensitivity and the prefill–decode balance. 
In the next section, we shift focus from workload properties to hardware-level
behavior and analyze how LLM inference energy and latency respond to GPU
frequency scaling under these heterogeneous workloads.

\section{DVFS Characterization}
\label{sec:dvfs_characterization}

This section characterizes the impact of GPU dynamic voltage and frequency
scaling (DVFS) on LLM inference energy and performance. Our analysis reveals a fundamental phase-level asymmetry: the decode phase, which dominates inference time, is memory-bound and largely insensitive to GPU frequency, enabling substantial energy savings with minimal latency penalty. We quantify this behavior across five models (1B--32B parameters), four
datasets, and seven discrete GPU SM frequency settings
(180, 487, 960, 1500, 2000, 2505 and 2842\,MHz), spanning the full supported operating range of the device.

\subsection{Experimental Setup}

This section builds on the experimental setup described in
Section~\ref{sec:experimental_setup}.
We evaluate five decoder-only LLMs (1B--32B parameters) across four datasets,
using fixed GPU SM frequencies spanning 180--2842\,MHz.
Energy and latency are measured at phase granularity (prefill and decode) under
batch sizes of 1, 4, and 8.
Unless stated otherwise, results are averaged over three runs with
1{,}000 queries per dataset.

  \begin{figure}[t]                                  
      \centering                                             
      \includegraphics[scale=0.28]{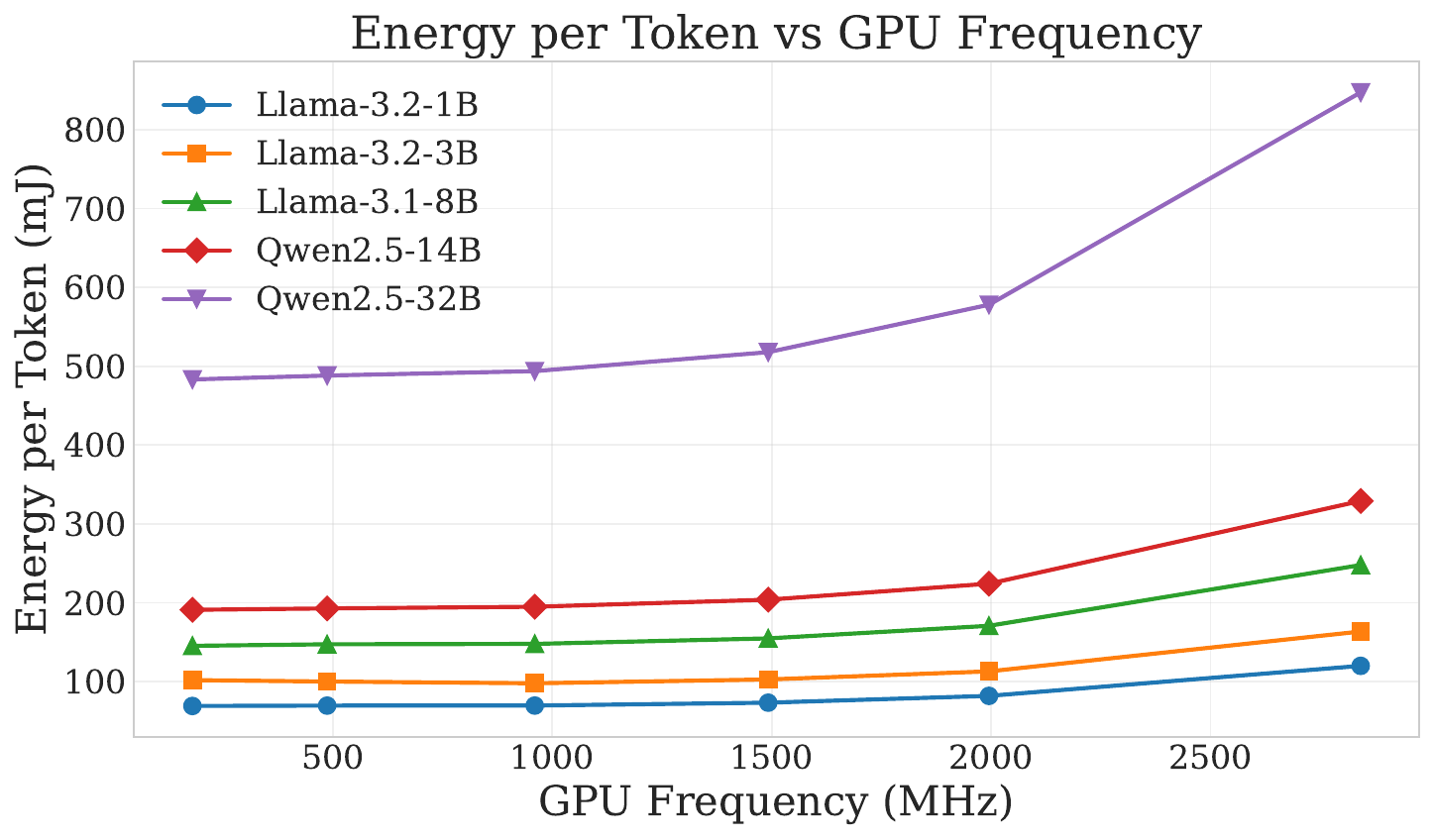}        
      \caption{Energy consumption per generated token across GPU frequencies. Lower frequencies achieve better energy efficiency (higher tokens per joule) due to the     
  memory-bound nature of the decode phase.}                                   
      \label{fig:energy-per-token}              
  \end{figure}  
\subsection{Energy Reduction via DVFS}
\label{sec:energy_reduction}

Figure~\ref{fig:energy-per-token} shows energy consumption per generated token as a
function of GPU SM frequency. Lowering frequency monotonically reduces energy consumption. At minimum frequency (180\,MHz), we observe \textbf{42\% average energy
savings} relative to baseline, ranging from 39.9\% (Llama-3.2-3B) to 44.2\%
(Llama-3.2-1B at batch 8). Table~\ref{tab:main_results} shows results across three batch sizes. Energy
savings are remarkably consistent: \textbf{41.9--43.6\%} across all batch sizes.
Latency increases are small: 2.8\% at batch 1, decreasing to just 1.1\% at batch 8.
Critically, decode latency changes are \textbf{negligible} ($\pm$1\%) across all
configurations, confirming decode's memory-bound nature. Prefill slowdown
decreases with batch size (25.7\% $\rightarrow$ 7.1\%) as larger batches amortize
the compute-bound prefill overhead. Larger models show smaller latency penalties
due to higher memory-boundedness.
This favorable tradeoff provides substantial energy savings with minimal performance
penalty across all batch sizes motivates our investigation into why LLM
inference is so amenable to frequency scaling.

  \begin{table}[t]                                                                                                                                                        
  \centering                                                                                                                                                          
  \caption{DVFS Results at 180\,MHz vs.\ Baseline (2842\,MHz).}                                                                                                           
  \label{tab:main_results}                                                                                                                                                
  \footnotesize                                                                                                                                                           
  \setlength{\tabcolsep}{2pt}                                                                                                                                             
  \begin{tabular}{lc cccc cc}                                                                                                                                             
  \toprule                                                                                                                                                                
  \textbf{Model} & \textbf{B} & \textbf{E}$\downarrow$ & \textbf{L}$\Delta$ & \textbf{Pre}$\Delta$ & \textbf{Dec}$\Delta$ & \textbf{Pre\%} & \textbf{Dec\%} \\            
  \midrule                                                                                                                                                                
              & 1 & 42.3\% & +5.6\%  & +52.4\% & +0.7\%   &  9.6\% & 90.4\% \\                                                                                            
  Llama-1B    & 4 & 42.4\% & +5.3\%  & +34.5\% & +1.2\%   & 12.4\% & 87.6\% \\                                                                                            
              & 8 & 44.2\% & +3.5\%  & +21.7\% & $-$0.3\% & 17.0\% & 83.0\% \\                                                                                            
  \midrule                                                                                                                                                                
              & 1 & 39.9\% & +4.2\%  & +37.4\% & +0.8\%   &  9.4\% & 90.6\% \\                                                                                            
  Llama-3B    & 4 & 42.3\% & +3.3\%  & +16.1\% & +1.0\%   & 14.7\% & 85.3\% \\                                                                                            
              & 8 & 43.5\% & +1.6\%  & +8.5\%  & $-$0.1\% & 20.0\% & 80.0\% \\                                                                                            
  \midrule                                                                                                                                                                
              & 1 & 42.2\% & +2.5\%  & +23.3\% & +0.1\%   & 10.0\% & 90.0\% \\                                                                                            
  Llama-8B    & 4 & 42.4\% & +0.8\%  & +7.6\%  & $-$0.5\% & 16.9\% & 83.1\% \\                                                                                            
              & 8 & 42.9\% & +0.3\%  & +3.4\%  & $-$0.7\% & 24.4\% & 75.6\% \\                                                                                            
  \midrule                                                                                                                                                                
              & 1 & 41.0\% & +1.3\%  & +11.2\% & $-$0.2\% & 13.8\% & 86.2\% \\                                                                                            
  Qwen-14B    & 4 & 41.7\% & +0.8\%  & +3.4\%  & +0.2\%   & 17.6\% & 82.4\% \\                                                                                            
              & 8 & 43.3\% & +0.0\%  & +1.4\%  & $-$0.5\% & 25.9\% & 74.1\% \\                                                                                            
  \midrule                                                                                                                                                                
              & 1 & 44.0\% & +0.3\%  & +4.4\%  & $-$0.2\% & 11.5\% & 88.5\% \\                                                                                            
  Qwen-32B    & 4 & 43.0\% & +0.3\%  & +1.7\%  & +0.0\%   & 19.2\% & 80.8\% \\                                                                                            
              & 8 & 44.1\% & $-$0.1\% & +0.5\% & $-$0.3\% & 27.5\% & 72.5\% \\                                                                                            
  \midrule                                                                                                                                                                
  \textbf{Avg B=1} & & \textbf{41.9\%} & \textbf{+2.8\%} & \textbf{+25.7\%} & \textbf{+0.2\%} & \textbf{10.9\%} & \textbf{89.1\%} \\                                      
  \textbf{Avg B=4} & & \textbf{42.4\%} & \textbf{+2.1\%} & \textbf{+12.7\%} & \textbf{+0.4\%} & \textbf{16.2\%} & \textbf{83.8\%} \\                                      
  \textbf{Avg B=8} & & \textbf{43.6\%} & \textbf{+1.1\%} & \textbf{+7.1\%}  & \textbf{$-$0.4\%} & \textbf{23.0\%} & \textbf{77.0\%} \\                                    
  \bottomrule                                                                                                                                                             
  \end{tabular}                                                              \par\vspace{2pt}
\noindent{\footnotesize
\textit{Notes:}
B = batch size;
E$\downarrow$ = energy reduction vs.\ baseline;
L$\Delta$ = end-to-end latency change;
Pre$\Delta$ = prefill latency change;
Dec$\Delta$ = decode latency change;
Pre\%, Dec\% = fraction of inference time.}
\end{table}      


\subsection{Why DVFS Works: Phase-Level Analysis}
\label{sec:phase_analysis}

The key to understanding DVFS effectiveness lies in the distinct behavior of
the two inference phases. Table~\ref{tab:main_results} 
reveals a striking asymmetry:

  \begin{itemize}             
  \item \textbf{Prefill latency increases by 7--26\%} at minimum frequency,                    
  with smaller increases at larger batch sizes, indicating compute-bound behavior.                   
  \item \textbf{Decode latency changes by $<$1\%}, indicating memory-bound                                                                              
  behavior largely insensitive to frequency.             
  \end{itemize}                                                         
This asymmetry arises from fundamental differences in how each phase utilizes
hardware. Prefill processes all input tokens in parallel through dense matrix
multiplications, achieving high arithmetic intensity that benefits from faster
clocks. Decode generates tokens one at a time and requires repeated memory accesses to the KV cache and model weights. As a result, it is memory-bandwidth-limited and largely insensitive to compute frequency.

Table~\ref{tab:main_results} (rightmost columns) quantifies phase contribution: at batch size 1, decode accounts for \textbf{86--91\% of total time}. As batch size increases, prefill's share grows: at batch 8,
prefill contributes 17--28\% of time (vs 9--14\% at batch 1). Decode throughput
(tokens/second) remains constant across all frequencies, while power consumption
increases significantly at maximum frequency. Higher frequencies during the decode phase increase energy consumption without providing measurable performance benefits.

\subsection{The Frequency Sweet Spot}
\label{sec:sweet_spot}

While minimum frequency maximizes energy savings, it incurs prefill slowdown.
We identify the optimal operating point by analyzing the energy-delay product
(EDP = Energy $\times$ Latency) across frequencies.

 \begin{table}[t]                                                                                                                                                        
  \centering                                                                                                                                                              
  \caption{Optimal EDP Frequency by Model and Batch Size (vs.\ 2842\,MHz).}                                                                                               
  \label{tab:optimal_freq}                                                                                                                                                
  \footnotesize                                                                                                                                                           
  \setlength{\tabcolsep}{2pt}                                                                                                                                             
  \begin{tabular}{l|ccc|ccc|ccc}                                                                                                                                          
  \toprule                                                                                                                                                                
   & \multicolumn{3}{c|}{\textbf{B=1}} & \multicolumn{3}{c|}{\textbf{B=4}} & \multicolumn{3}{c}{\textbf{B=8}} \\                                                          
  \textbf{Model} & \textbf{Freq} & \textbf{E}$\downarrow$ & \textbf{L} & \textbf{Freq} & \textbf{E}$\downarrow$ & \textbf{L} & \textbf{Freq} & \textbf{E}$\downarrow$ &   
  \textbf{L} \\                                                                                                                                                           
  \midrule                                                                                                                                                                
  Llama-1B  & 960 & 44.8\% & $-$4.4\%  & 960 & 42.3\% & +1.2\% & 180 & 44.2\% & +3.5\% \\                                                                                 
  Llama-3B  & 960 & 47.6\% & $-$10.4\% & 487 & 42.7\% & +2.7\% & 960 & 42.8\% & +0.2\% \\                                                                                 
  Llama-8B  & 960 & 47.4\% & $-$11.0\% & 180 & 42.4\% & +0.8\% & 960 & 45.2\% & $-$5.6\% \\                                                                               
  Qwen-14B  & 960 & 51.5\% & $-$19.0\% & 487 & 41.8\% & +0.5\% & 180 & 43.3\% & +0.0\% \\                                                                                 
  Qwen-32B  & 960 & 63.2\% & $-$36.5\% & 487 & 43.4\% & +0.2\% & 180 & 44.1\% & $-$0.1\% \\                                                                               
  \bottomrule                                                                                                                                                             
  \end{tabular}                                                               \par\vspace{2pt}
\noindent{\footnotesize
\textit{Notes:}
Freq = SM frequency (MHz) minimizing EDP;
E$\downarrow$ = energy reduction vs.\ 2842\,MHz;
L = end-to-end latency change.}
 
  \end{table}

\begin{figure}[t]
    \centering
    \includegraphics[scale=0.25]{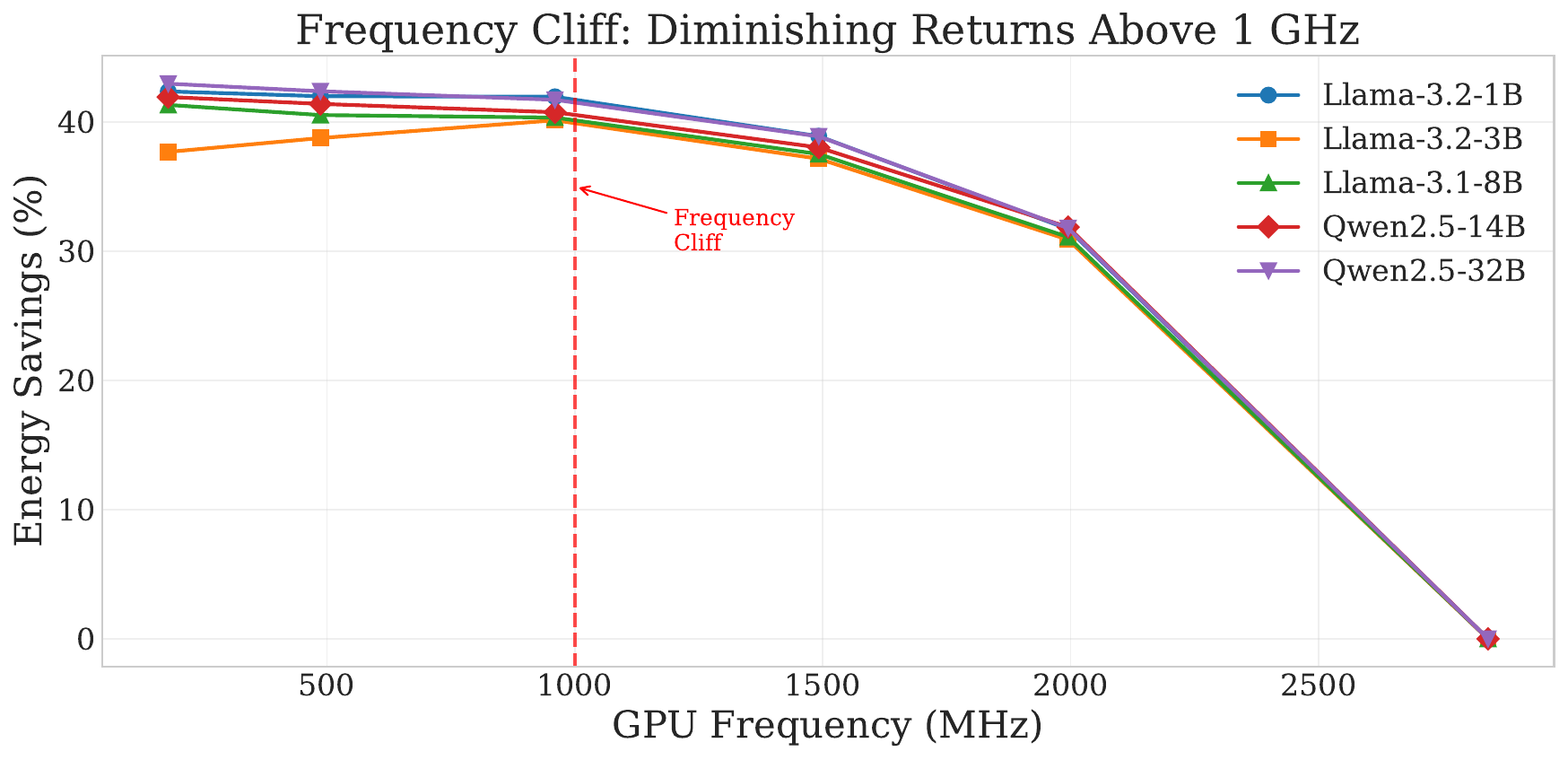}
    \caption{The frequency cliff: energy savings plateau below $\sim$1000\,MHz. All models achieve 40--45\% savings in the plateau region, with diminishing returns at lower frequencies. The optimal operating point lies at $\sim$960\,MHz where energy savings are maximized without significant latency penalty.}
    \label{fig:frequency_cliff}
\end{figure}

Table~\ref{tab:optimal_freq} shows optimal EDP frequencies across batch sizes.
At batch size 1, all models converge to approximately \textbf{960\,MHz}
(34\% of maximum), delivering 44--63\% energy savings with latency improvements.
At larger batch sizes, optimal frequencies vary more across models, with some
achieving even greater savings. Figure~\ref{fig:frequency_cliff} visualizes
the frequency cliff phenomenon across all evaluated models. The sweet spot exists because:

\begin{enumerate}
\item Energy savings plateau below $\sim$1000\,MHz (the ``frequency cliff''),
yielding diminishing returns at lower frequencies.
\item Prefill slowdown accelerates below 1000\,MHz, increasing latency
disproportionately.
\end{enumerate}

\subsection{Effect of Output Length and Model Size}
\label{sec:output_length}

DVFS effectiveness depends on output length and model size, which determine decode's
contribution to total latency. Table~\ref{tab:workload_effects} shows these effects. For short-output datasets (BoolQ, HellaSwag), prefill accounts for 31--39\% of
time, resulting in latency increases of 4--8\%. For long-output datasets
(TruthfulQA, NarrativeQA), prefill drops to 3--12\%, and DVFS achieves
\textbf{42--43\% energy savings with under 1\% latency penalty}.
Model size affects latency impact: \textbf{small models (1--3B)} show 2--5\%
latency increase, \textbf{medium models (8B)} achieve 0.3--2.5\% increase, and
\textbf{large models (14--32B)} show negligible impact ($<$1\%). All categories
remain memory-bound with consistent 41--44\% energy savings.

\begin{table}[t]
\centering
\caption{DVFS Effectiveness by Output Length and Model Size (180\,MHz, B=1).}
\label{tab:workload_effects}
\footnotesize
\setlength{\tabcolsep}{2pt}
\begin{tabular}{lcc|lcc}
\toprule
\textbf{Dataset} & \textbf{E$\downarrow$} & \textbf{L$\uparrow$} & \textbf{Model} & \textbf{E$\downarrow$} & \textbf{L$\uparrow$} \\
\midrule
BoolQ (LL)     & 41.4\% & +7.7\% & Small (1--3B)  & 40.9\% & +4.8\% \\
HellaSwag (LL) & 42.6\% & +4.1\% & Medium (8B)    & 42.2\% & +2.5\% \\
TruthfulQA (100)   & 42.5\% & +0.7\% & Large (14--32B)& 43.2\% & +0.6\% \\
NarrativeQA (100)  & 43.1\% & +0.9\% & & & \\
\bottomrule
\end{tabular}
\par\vspace{2pt}
  {\footnotesize \textit{LL = log-likelihood evaluation (no token generation).}}    
\end{table}

\subsection{Effect of Batch Size}
\label{sec:batch_size}

\begin{figure}[t]
    \centering
    \includegraphics[width=\columnwidth]{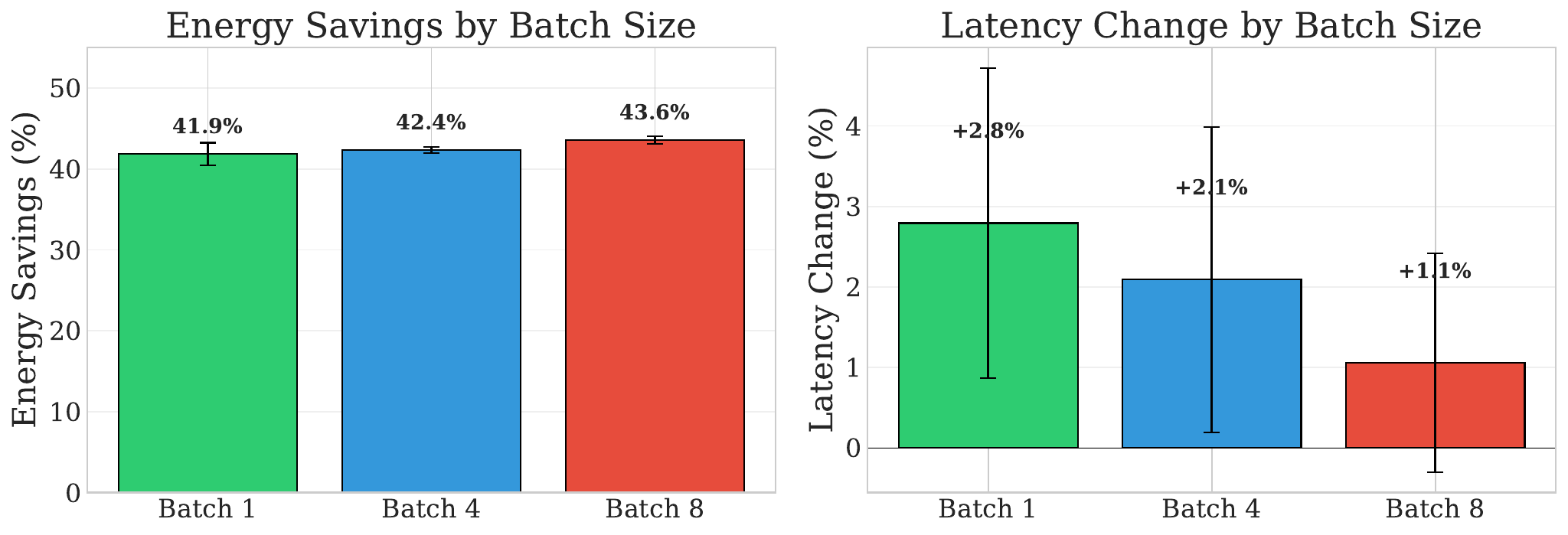}
    \caption{Effect of batch size on DVFS effectiveness. Energy savings remain consistent (42--44\%) across all batch sizes. Latency penalties decrease with larger batches as prefill overhead is amortized over more tokens.}
    \label{fig:batch_size}
\end{figure}

Contrary to initial expectations, batch size has minimal impact on DVFS
effectiveness. As shown in Table~\ref{tab:main_results}, energy savings remain
consistent at \textbf{41.9--43.6\%} across batch sizes 1, 4, and 8
(Figure~\ref{fig:batch_size}). Latency impact decreases with larger batches, from
+2.8\% at batch 1 to +1.1\% at batch 8, as prefill overhead is amortized.
This robustness arises because the decode phase remains memory-bound even at larger batch sizes. While prefill becomes more compute-intensive with batching, the decode
phase that dominates total time remains bottlenecked by memory bandwidth for
loading model weights and the KV cache. DVFS provides \textbf{consistent
42\% energy savings} with modest latency impact (1--3\%) across batch sizes,
making low-frequency operation suitable for offline and latency-tolerant
scenarios.

\subsection{Summary and Recommendations}
\label{sec:recommendations}

  \begin{table}[t]                                 
  \centering                                     
  \caption{Summary of phase-level DVFS effects on energy and latency.}                                   
  \label{tab:summary}                                         
  \footnotesize                                     
  \setlength{\tabcolsep}{4pt}                                   
  \begin{tabular}{l c}                                       
  \toprule                                         
  \textbf{Aspect} & \textbf{Observation} \\                                        
  \midrule                                       
  \multicolumn{2}{l}{\textit{DVFS at 180\,MHz (vs.\ 2842\,MHz)}} \\                                                                                     
  Energy savings            & 40--44\% (avg.\ 42\%) \\                                           
  Latency change            & +1--3\% \\                                                        
  Prefill / Decode slowdown & +7--26\% / $\pm$1\% \\                                          
  Decode time fraction      & 77--91\% \\                                                       
  \midrule                                                                                  
  \multicolumn{2}{l}{\textit{Effect of Batch Size (B=1/4/8)}} \\                                                                                        
  Energy savings            & 41.9 / 42.4 / 43.6\% \\                                       
  Latency impact            & +2.8 / +2.1 / +1.1\% \\                                                      
  \midrule                                      
  \multicolumn{2}{l}{\textit{Operating Point Tradeoffs}} \\                                                                                             
  Max savings (180\,MHz)     & 42\% savings, +1--3\% latency \\                                                                                         
  EDP-optimal ($\sim$960\,MHz) & 45--50\% savings, $-$5--10\% latency \\          
  Baseline (2842\,MHz)       & Reference point \\                                   
  \bottomrule                                       
  \end{tabular}                                         \end{table}

Table~\ref{tab:summary} consolidates our key findings.
The central insight is that \textbf{LLM inference is dominated by memory-bound
decode operations that are largely insensitive to GPU frequency}, enabling
\textbf{42\% energy savings with only 1--3\% latency increase}. While aggressive
DVFS (down to 180\,MHz) proved effective in our controlled setting, these results
motivate \emph{phase-aware} GPU frequency policies rather than a universal
operating point.

\section{Case Study: Implications of Workload-Aware DVFS}
\label{sec:casestudy}

To illustrate the practical implications of our workload characterization and DVFS findings, we present a case study exploring how query difficulty features could inform energy optimization decisions. This analysis examines the potential synergy between model routing and frequency scaling and provides empirical estimates of achievable energy savings.

\subsection{Mapping Query Difficulty to Model Scale}
\label{sec:casestudy:routing}

Our scaling pattern analysis (Section~\ref{sec:workload:difficulty}) revealed that query difficulty varies systematically with input features. We explore how these patterns might inform model selection decisions.
The four scaling patterns identified in our workload characterization suggest a natural mapping to model tiers. Queries classified as ``Always Easy'' (44.5\% of our dataset) achieved comparable quality scores across all model sizes, suggesting they could potentially be handled by smaller models. Conversely, ``Scaling Helps'' queries (15.5\%) showed measurable quality improvements with larger models, while ``Always Hard'' queries (32.6\%) exhibited limited benefit from increased model capacity. Table~\ref{tab:routing_strategy} summarizes how these patterns could map to routing decisions.

  \begin{table}[t]
  \centering
  \caption{Routing Strategy Based on Scaling Patterns.}
  \label{tab:routing_strategy}
  \footnotesize
  \setlength{\tabcolsep}{3pt}
  \begin{tabular}{lccc}
  \toprule
  \textbf{Pattern} & \textbf{\%} & \textbf{Model} & \textbf{Rationale} \\
  \midrule
  Always Easy   & 44.5 & 1--3B & Similar quality across sizes \\
  Scaling Helps & 15.5 & 8B+   & Quality improves with scale \\
  Always Hard   & 32.6 & 1--3B & Limited benefit from scaling \\
  Inconsistent  &  7.4 & 8B    & Architecture-dependent \\
  \bottomrule
  \end{tabular}
  \end{table}

The energy consumption measurements from Section~\ref{sec:dvfs_characterization} provide context for these routing decisions. At baseline frequency (2842~MHz), average energy per query ranged from 2.9~J for the 1B model to 21.0~J for the 32B model, a 7.2$\times$ difference. If easy queries were routed to 3B models instead of 32B, this would represent an 80\% energy reduction per query for that subset of the workload.



\subsection{Phase-Aware Frequency Scaling}
\label{sec:casestudy:dvfs}

The distinct behavior of prefill and decode phases
(Section~\ref{sec:dvfs_characterization}) motivates phase-aware frequency scaling. Since decode dominates inference time (77--91\%) and is largely memory-bound, we consider a policy that uses high frequency during prefill (2842\,MHz) and switches to low frequency during decode (180\,MHz). Across five models, this reduces energy consumption by 41.9\% on average with
a 2.8\% latency increase (Table \ref{tab:dvfs_per_model}). Larger models incur smaller penalties (0.3\% for 32B vs.\ 5.6\% for 1B), and long-output workloads achieve 43--44\% savings with
under 1\% latency impact. Results at 180\,MHz represent an upper bound; in practice, EDP-optimal frequencies as shown in Table~\ref{tab:optimal_freq} may be preferred.

\begin{figure}[t]
    \centering
    \includegraphics[scale=0.3]{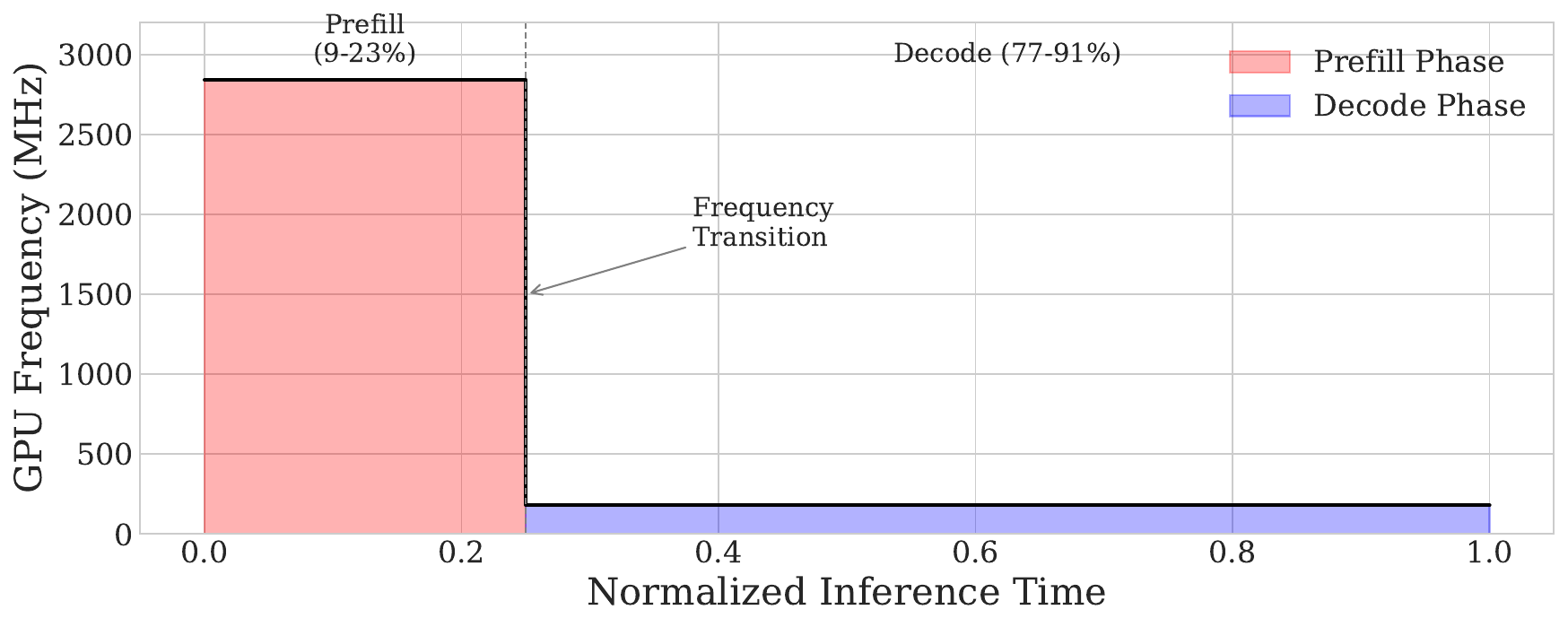}
    \caption{Phase-aware frequency profile during inference. The transition occurs after prefill completion, exploiting the memory-bound nature of decode.}
    \label{fig:phase_timeline}
\end{figure}

\begin{table}[t]
\centering
\caption{DVFS Energy Savings by Model (2842~MHz $\rightarrow$ 180~MHz)}
\label{tab:dvfs_per_model}
\begin{tabular}{lcccc}
\toprule
\textbf{Model} & \textbf{Baseline (J)} & \textbf{Low Freq (J)} & \textbf{Savings} & \textbf{Latency} \\
\midrule
Llama-3.2-1B & 2.92 & 1.69 & 42.3\% & +5.6\% \\
Llama-3.2-3B & 4.19 & 2.52 & 39.9\% & +4.2\% \\
Llama-3.1-8B & 6.24 & 3.61 & 42.2\% & +2.5\% \\
Qwen2.5-14B & 8.32 & 4.91 & 41.0\% & +1.3\% \\
Qwen2.5-32B & 20.97 & 11.74 & 44.0\% & +0.3\% \\
\midrule
\multicolumn{3}{l}{\textbf{Average}} & \textbf{41.9\%} & \textbf{+2.8\%} \\
\bottomrule
\end{tabular}
\end{table}

\subsection{Combined Optimization Potential}
\label{sec:casestudy:combined}

Combining model routing with phase-aware frequency scaling could yield multiplicative energy reductions. We estimate the potential savings by applying both techniques to our workload distribution. For each query category, we compute expected savings by combining the model routing benefit (relative to always using the 32B model at 2842~MHz) with the DVFS benefit for the target model tier. Table~\ref{tab:combined_estimate} presents these projections. These projections suggest that workload-aware optimization could reduce energy consumption by approximately 87\% compared to a baseline of always using the largest model at maximum frequency.

\begin{table}[t]
\centering
\caption{Estimated Combined Energy Savings}
\label{tab:combined_estimate}
\begin{tabular}{lcccc}
\toprule
\textbf{Category} & \textbf{\%} & \textbf{Model} & \textbf{Freq} & \textbf{Est. Savings} \\
\midrule
Always Easy & 44.5\% & 3B & 180~MHz & 88\% \\
Scaling Helps & 15.5\% & 14B & 180~MHz & 77\% \\
Always Hard & 32.6\% & 3B & 180~MHz & 88\% \\
Inconsistent & 7.4\% & 8B & 180~MHz & 83\% \\
\midrule
\multicolumn{4}{l}{\textbf{Weighted Average}} & \textbf{87\%} \\
\bottomrule
\end{tabular}
\end{table}



\subsubsection{Quality Considerations}

\begin{figure}[t]
    \centering
    \includegraphics[width=0.9\columnwidth]{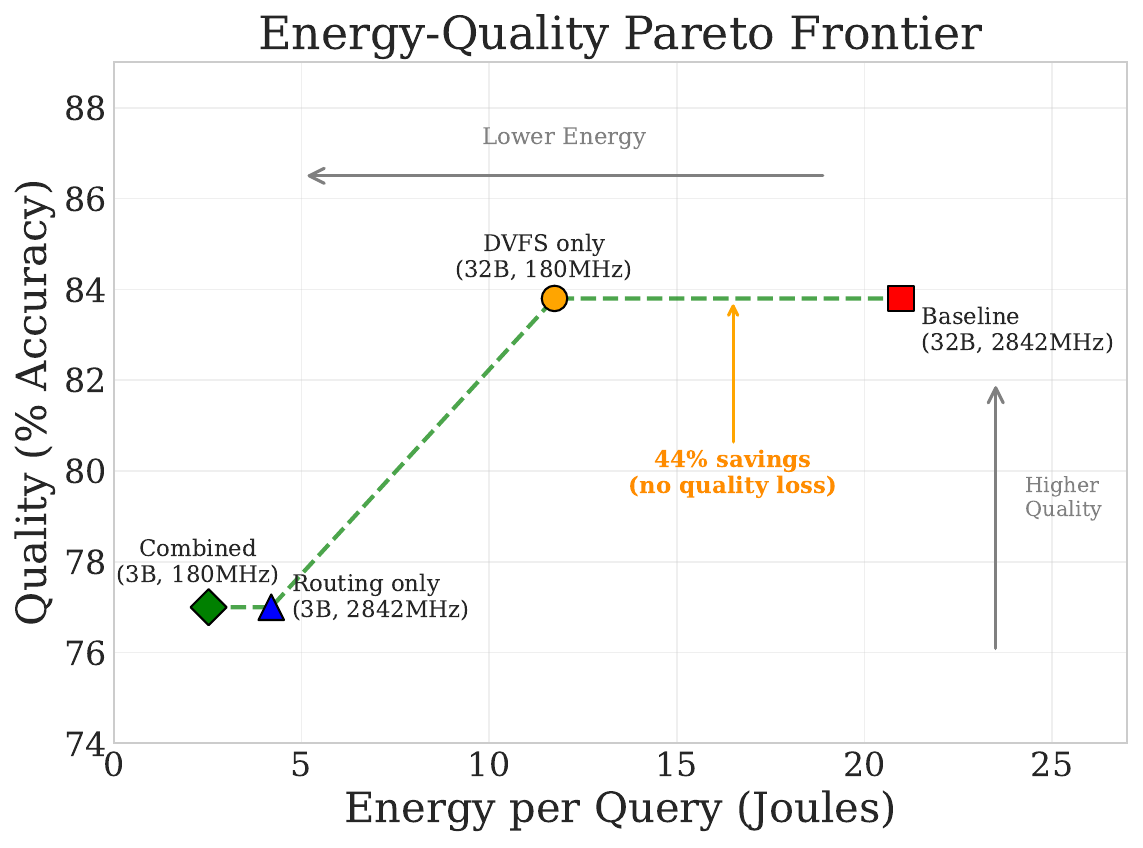}
    \caption{Energy-quality Pareto frontier. DVFS provides ``free'' energy savings; routing introduces quality tradeoffs for additional gains.}
    \label{fig:pareto}
\end{figure}

An important consideration is the quality impact of routing queries to smaller models. Our validation analysis found that overall classification accuracy (averaged across BoolQ and HellaSwag) was 77.0\% on the 3B model versus 83.8\% on the 32B model, a gap of 6.8 percentage points. Whether this tradeoff is acceptable depends on application requirements.
Table~\ref{tab:pareto} positions different optimization strategies on the energy-quality frontier. The combined approach achieves the highest energy savings but incurs a modest quality reduction concentrated in the subset of queries routed to smaller models. Figure~\ref{fig:pareto} visualizes these tradeoffs, showing that DVFS-only optimization preserves quality while routing introduces a quality-energy tradeoff.

\begin{table}[t]
\centering
\caption{Energy-Quality Tradeoff Across Strategies}
\label{tab:pareto}
\begin{tabular}{lccc}
\toprule
\textbf{Strategy} & \textbf{Energy} & \textbf{Quality} & \textbf{Est. Savings} \\
\midrule
Baseline (32B, 2842~MHz) & 20.97~J & 83.8\% & --- \\
DVFS only (32B, 180~MHz) & 11.74~J & 83.8\% & 44\% \\
Routing only (3B, 2842~MHz) & 4.19~J & 77.0\% & 80\% \\
Combined (3B, 180~MHz) & 2.52~J & 77.0\% & 88\% \\
\bottomrule
\end{tabular}
\end{table}

\subsection{Main Insights from the Case Study}

This case study illustrates how workload characterization findings could inform practical energy optimization. Three key observations emerge:

\begin{enumerate}
    \item \textbf{Model routing potential:} The finding that 44.5\% of queries achieve similar quality across model sizes suggests significant routing opportunities. Routing to 3B instead of 32B yields 80\% energy savings with a 6.8 percentage point quality reduction.

    \item \textbf{DVFS viability:} Reducing GPU frequency from 2842~MHz to 180~MHz achieves 42\% average energy savings with 3\% latency increase, preserving quality. Larger models incur smaller latency penalties due to their memory-bound nature.

    \item \textbf{Combined potential:} Workload-aware optimization combining both techniques could achieve up to 87\% energy reduction, though quality implications warrant application-specific evaluation.
\end{enumerate}

\noindent \textbf{Threats to validity.}
Our evaluation uses offline replay-based benchmarking on a single GPU and does
not capture production serving dynamics such as continuous batching,
speculative decoding, or multi-GPU execution. Energy measurements reflect GPU
consumption only and exclude system-level overheads. Extending this analysis to
distributed inference, heterogeneous hardware, and non-English workloads is an
important direction for future work. Moreover, 
these findings are intended as empirical observations rather than deployment prescriptions. Real-world implementation would require additional considerations, including classification accuracy, switching overhead, and service-level requirements.

\section{Conclusion}
\label{sec:conclusion}

We present a measurement-driven characterization of LLM inference energy efficiency across workload and hardware dimensions. We showed that semantic query features predict inference difficulty better than input length, with 44.5\% of queries achieving comparable quality across model sizes, and that the decode phase dominates inference time (77--91\%) while remaining largely insensitive to GPU frequency. This enables up to 42\% energy savings at 180\,MHz with only 1--3\% latency increase. These findings challenge the assumptions that inference cost scales primarily with token count or requires a uniform GPU frequency. A case study further illustrated that combining workload-aware model selection with DVFS can reduce energy consumption by up to 87\%, providing an empirical foundation for phase-aware and workload-aware optimization in future LLM serving systems.


Future work will explore phase-aware runtime DVFS control, lightweight query classification for model routing, and extensions to multi-GPU inference and emerging accelerator architectures.

 \section*{Acknowledgment}


This work was partially supported by the Austrian Science Fund (FWF) through the Themis project “Trustworthy and Sustainable Code Offloading” (Grant DOI: 10.55776/PAT1668223) and the Standalone Project “Transprecise Edge Computing (Triton)” (Grant No. P 36870-N). It was further supported by the Austrian Research Promotion Agency (FFG) through the Virtual Shepherd project (Grant No. FO999910627) and by the Vienna Science and Technology Fund (WWTF) through the AI-supported Holographic Environmental Water Monitoring project (Grant No. ESR24-053).

\bibliographystyle{IEEEtran}
\bibliography{references}
\end{document}